%% file: main.tex
\documentclass{article}

\PassOptionsToPackage{numbers}{natbib}
\usepackage[preprint]{neurips_2026}

\input{preamble}

\usepackage[utf8]{inputenc} %
\usepackage[T1]{fontenc}    %
\usepackage{hyperref}       %
\usepackage{cleveref}
\crefname{figure}{Fig.}{Figs.}
\crefname{section}{Sec.}{Secs.}
\crefname{table}{Tab.}{Tabs.}

\title{Self-Supervised Learning with a Multi-Task Latent~Space~Objective}

\author{%
  \textbf{Pierre-François De Plaen}\textsuperscript{1} \quad \textbf{Abhishek Jha}\textsuperscript{2,*} \quad \textbf{Luc Van Gool}\textsuperscript{1,3,4,5} 
  \\ \textbf{Tinne Tuytelaars}\textsuperscript{1} \quad \textbf{Marc Proesmans}\textsuperscript{1,5} 
  \\
    \textsuperscript{1}ESAT-PSI, KU Leuven, Belgium \quad 
    \textsuperscript{2}VIB.AI, KU Leuven, Belgium \\ %
    \textsuperscript{3}CVL, ETH Zürich, Switzerland \quad
    \textsuperscript{4}INSAIT, Sofia University, Bulgaria \quad
    \textsuperscript{5}TRACE vzw 
}

\begin{document}

\maketitle

\begin{abstract}
  We propose a multi-task formulation of self-predictive Siamese SSL in which each spatial transformation defines a distinct latent-space alignment task, solved by a dedicated predictor over a shared encoder. This perspective directly explains a long-standing failure of multi-crop training in self-predictive methods such as BYOL, SimSiam, and MoCo v3: a shared predictor is forced to solve heterogeneous alignment tasks simultaneously, leading to unstable optimization. Assigning one predictor per view type resolves this interference, unlocking linear evaluation gains of 3.8-4\% across frameworks. 
  This perspective also suggests a principled way to enrich pre-training by introducing additional spatial transformations as complementary tasks. We demonstrate this by introducing asymmetric cutout views, in which a masked online view is aligned with a complete target, forming a semantic inpainting objective. The resulting framework is stable, backbone-agnostic, and consistently improves the performance of ResNet and ViT models on ImageNet and COCO.
\end{abstract}

\blfootnote{Code: \url{https://github.com/pfdp0/mulan}}
\blfootnote{*Part of this work was conducted while Abhishek was affiliated with ESAT-PSI, KU Leuven, Belgium.}

\input{sections/intro}
\input{sections/related}
\input{sections/method}
\input{sections/experiments}
\input{sections/conclusion}

\input{sections/acknowledgments}

\bibliographystyle{ieeenat_fullname}
\bibliography{bib}

\clearpage
\appendix
\input{sections/suppl}

\end{document}

%% file: sections/intro.tex
\section{Introduction}
Self-supervised learning (SSL) has become a dominant paradigm for learning visual representations without supervision. It builds training signals directly from data, allowing models to discover visual structure on their own. Among SSL approaches, Siamese-based methods, often referred to as Joint Embedding Architectures~\cite{chen2020simclr,he2020mocov1,grill2020byol,zbontar2021barlow,caron2021dino}, have proven especially effective. They learn by aligning representations of different augmented views of the same image. Recent large-scale efforts~\cite{oquab2023dinov2,simeoni2025DINOv3} have further advanced this paradigm by leveraging larger datasets and more powerful architectures, yielding representations that generalize remarkably across a wide range of tasks. 

An important factor behind this progress is the multi-crop strategy, which adds several small local crops to the pair of global views. This simple idea promotes spatial consistency and has been crucial to the success of clustering-based methods~\cite{caron2020swav,caron2021dino}. Yet, it turns out to be unstable in self-predictive Siamese architectures, such as BYOL~\cite{grill2020byol} and SimSiam~\cite{chen2021simsiam}, where the online and target branches play different roles, with the former including a prediction head. This instability has prevented these otherwise strong frameworks from benefiting from one of the most effective SSL augmentations.

\begin{figure}[t]%
    \centering
    \begin{subfigure}[T]{0.32\textwidth}%
        \centering
        \def\svgwidth{0.98\linewidth}
        {\scriptsize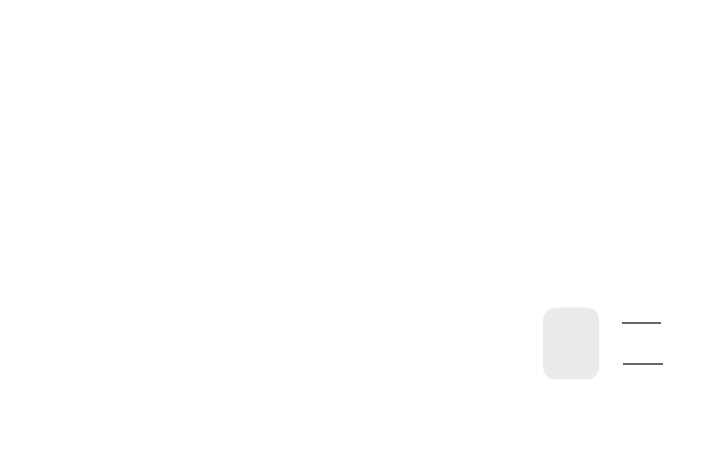}
        \caption{Naive multi-crop.}
        \label{subfig:teaser_naive_mc}
    \end{subfigure}
    \hfill
    \begin{subfigure}[T]{0.32\textwidth}%
        \centering
        \def\svgwidth{0.98\textwidth}
        {\scriptsize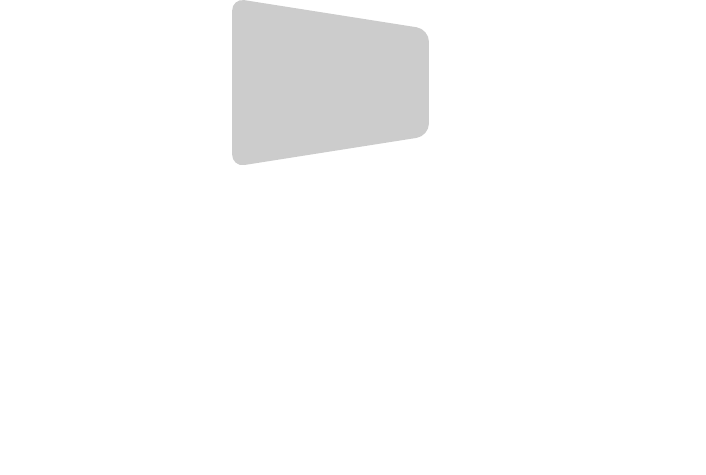}
        \caption{Multi-predictor multi-crop (ours, \Cref{sec:stabilized_multi_crop}).}
        \label{subfig:teaser_mp_mc}
    \end{subfigure}
    \hfill
    \begin{subfigure}[T]{0.32\textwidth}%
        \centering
        \def\svgwidth{0.98\textwidth}
        {\scriptsize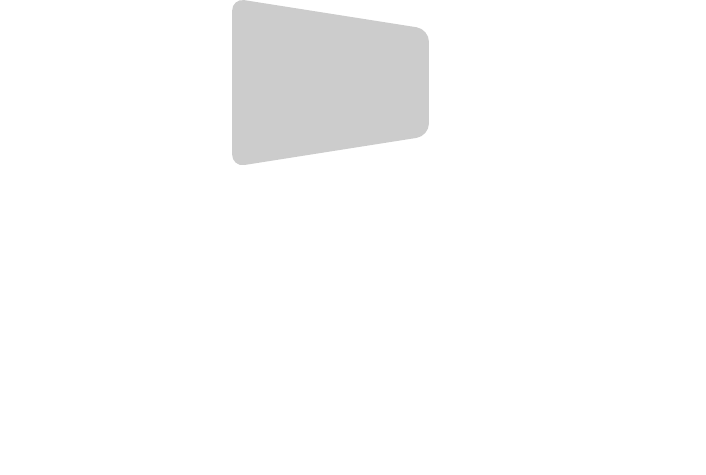}
        \caption{Multi-task (ours, \Cref{sec:multi_task_formulation}).}
        \label{subfig:teaser_mt}
    \end{subfigure}
    \caption{Overview of the proposed framework. Naive multi-crop (left) forces a single predictor to solve heterogeneous latent-space tasks simultaneously, leading to unstable optimization. Assigning one predictor per view type (middle) resolves this interference and stabilizes training. Introducing asymmetric cutout views (right) adds a complementary semantic inpainting task, further improving downstream performance. Here, \textit{p.} denotes a predictor, and colors indicate distinct predictors.}
    \label{fig:teaser}
\end{figure}

We analyze this limitation and trace it to the shared predictor used across all views. A single predictor must align representations from both global and local crops, which differ strongly in scale and content, leading to unstable optimization.
We resolve this by assigning a dedicated predictor to each view type, while keeping the encoder shared (\Cref{subfig:teaser_mp_mc}).
This simple modification stabilizes multi-crop training, yielding consistent accuracy gains across frameworks. 
On ImageNet, it improves linear evaluation by 3.8-4\% across BYOL, SimSiam, and MoCo v3 (\Cref{tab:multi-crop_rocks}).

This multi-predictor design suggests a broader perspective: each spatial transformation defines a distinct pre-training task, and new transformations can be introduced as complementary tasks without modifying the architecture or loss. For example, local crops define a local-to-global alignment task, encouraging the model to infer global context from partial observations. This raises a natural question: which additional transformations induce useful complementary tasks?

We study supervisory signals in SSL (\Cref{sec:spatial_augs_study}) and find that spatial augmentations are the primary driver of representation learning. Motivated by this, we propose \emph{asymmetric cutout}~\cite{devries2017cutout}: a region is masked in the online view while the target remains unaltered, creating a semantic inpainting objective in latent space. We first validate it as a standalone task and find that asymmetry is crucial, as masking both views causes performance to collapse, confirming that the model must predict the complete target representation from a masked view (\Cref{sec:assym_cutout_validation}).

Together, these elements form a unified and stable framework for self-predictive SSL (\Cref{subfig:teaser_mt}). Each spatial transformation defines a distinct latent-space task with its own predictor. All tasks share a single encoder (backbone and projection head) and the same alignment loss, yielding a simple multi-task formulation compatible with both CNNs and transformers.

In summary, our contributions are the following:
\begin{enumerate}[leftmargin=2em,nosep]
    \item We identify the shared predictor as the primary cause of multi-crop instability in self-predictive SSL methods and show that decoupling predictors by view-type enables these methods to benefit from the multi-crop strategy.
    \item We study supervisory signals in SSL and confirm the central role of spatial transformations. We further provide evidence that multi-crop gains stem from increased spatial diversity rather than specifically from the inclusion of low-resolution crops.
    \item We interpret spatial transformations as latent-space tasks and extend our framework with a semantic inpainting task, which further improves performance. 
    \item We validate the approach on key self-predictive methods with both ResNet and ViT backbones, achieving faster convergence and consistent performance gains across ImageNet (kNN, linear, semi-supervised, and fine-tuning) and COCO (detection and segmentation).
\end{enumerate}

%% file: figures/teaser_fig_a.pdf_tex
\begingroup%
  \makeatletter%
  \providecommand\color[2][]{%
    \errmessage{(Inkscape) Color is used for the text in Inkscape, but the package 'color.sty' is not loaded}%
    \renewcommand\color[2][]{}%
  }%
  \providecommand\transparent[1]{%
    \errmessage{(Inkscape) Transparency is used (non-zero) for the text in Inkscape, but the package 'transparent.sty' is not loaded}%
    \renewcommand\transparent[1]{}%
  }%
  \providecommand\rotatebox[2]{#2}%
  \newcommand*\fsize{\dimexpr\f@size pt\relax}%
  \newcommand*\lineheight[1]{\fontsize{\fsize}{#1\fsize}\selectfont}%
  \ifx\svgwidth\undefined%
    \setlength{\unitlength}{338.57238264bp}%
    \ifx\svgscale\undefined%
      \relax%
    \else%
      \setlength{\unitlength}{\unitlength * \real{\svgscale}}%
    \fi%
  \else%
    \setlength{\unitlength}{\svgwidth}%
  \fi%
  \global\let\svgwidth\undefined%
  \global\let\svgscale\undefined%
  \makeatother%
  \begin{picture}(1,0.67309487)%
    \lineheight{1}%
    \setlength\tabcolsep{0pt}%
    \put(0,0){\includegraphics[width=\unitlength,page=1]{teaser_fig_a.pdf}}%
    \put(0.79261555,0.17499951){\color[rgb]{0,0,0}\makebox(0,0)[lt]{\lineheight{1.25}\smash{\begin{tabular}[t]{l}p.\end{tabular}}}}%
    \put(0,0){\includegraphics[width=\unitlength,page=2]{teaser_fig_a.pdf}}%
    \put(0.37695849,0.54271999){\color[rgb]{0,0,0}\makebox(0,0)[lt]{\lineheight{1.25}\smash{\begin{tabular}[t]{l}Encoder\end{tabular}}}}%
    \put(0.00015576,0.54271999){\color[rgb]{0,0,0}\makebox(0,0)[lt]{\lineheight{1.25}\smash{\begin{tabular}[t]{l}Targets\end{tabular}}}}%
    \put(0,0){\includegraphics[width=\unitlength,page=3]{teaser_fig_a.pdf}}%
    \put(0.8903273,0.41665389){\color[rgb]{0,0,0}\makebox(0,0)[lt]{\lineheight{1.25}\smash{\begin{tabular}[t]{l}Loss\end{tabular}}}}%
    \put(0,0){\includegraphics[width=\unitlength,page=4]{teaser_fig_a.pdf}}%
    \put(0.37695849,0.17499951){\color[rgb]{0,0,0}\makebox(0,0)[lt]{\lineheight{1.25}\smash{\begin{tabular}[t]{l}Encoder\end{tabular}}}}%
    \put(0.016969,0.23452149){\color[rgb]{0,0,0}\makebox(0,0)[lt]{\lineheight{1.25}\smash{\begin{tabular}[t]{l}Global\end{tabular}}}}%
    \put(0.0459436,0.11490157){\color[rgb]{0,0,0}\makebox(0,0)[lt]{\lineheight{1.25}\smash{\begin{tabular}[t]{l}Local\end{tabular}}}}%
    \put(0,0){\includegraphics[width=\unitlength,page=5]{teaser_fig_a.pdf}}%
  \end{picture}%
\endgroup%

%% file: figures/teaser_fig_b.pdf_tex
\begingroup%
  \makeatletter%
  \providecommand\color[2][]{%
    \errmessage{(Inkscape) Color is used for the text in Inkscape, but the package 'color.sty' is not loaded}%
    \renewcommand\color[2][]{}%
  }%
  \providecommand\transparent[1]{%
    \errmessage{(Inkscape) Transparency is used (non-zero) for the text in Inkscape, but the package 'transparent.sty' is not loaded}%
    \renewcommand\transparent[1]{}%
  }%
  \providecommand\rotatebox[2]{#2}%
  \newcommand*\fsize{\dimexpr\f@size pt\relax}%
  \newcommand*\lineheight[1]{\fontsize{\fsize}{#1\fsize}\selectfont}%
  \ifx\svgwidth\undefined%
    \setlength{\unitlength}{338.57238264bp}%
    \ifx\svgscale\undefined%
      \relax%
    \else%
      \setlength{\unitlength}{\unitlength * \real{\svgscale}}%
    \fi%
  \else%
    \setlength{\unitlength}{\svgwidth}%
  \fi%
  \global\let\svgwidth\undefined%
  \global\let\svgscale\undefined%
  \makeatother%
  \begin{picture}(1,0.67309487)%
    \lineheight{1}%
    \setlength\tabcolsep{0pt}%
    \put(0,0){\includegraphics[width=\unitlength,page=1]{teaser_fig_b.pdf}}%
    \put(0.37695849,0.54271999){\color[rgb]{0,0,0}\makebox(0,0)[lt]{\lineheight{1.25}\smash{\begin{tabular}[t]{l}Encoder\end{tabular}}}}%
    \put(0.00015576,0.54271999){\color[rgb]{0,0,0}\makebox(0,0)[lt]{\lineheight{1.25}\smash{\begin{tabular}[t]{l}Targets\end{tabular}}}}%
    \put(0,0){\includegraphics[width=\unitlength,page=2]{teaser_fig_b.pdf}}%
    \put(0.8903273,0.41665389){\color[rgb]{0,0,0}\makebox(0,0)[lt]{\lineheight{1.25}\smash{\begin{tabular}[t]{l}Loss\end{tabular}}}}%
    \put(0,0){\includegraphics[width=\unitlength,page=3]{teaser_fig_b.pdf}}%
    \put(0.37695849,0.17499951){\color[rgb]{0,0,0}\makebox(0,0)[lt]{\lineheight{1.25}\smash{\begin{tabular}[t]{l}Encoder\end{tabular}}}}%
    \put(0.016969,0.23452149){\color[rgb]{0.19607843,0.52941176,0.69803922}\makebox(0,0)[lt]{\lineheight{1.25}\smash{\begin{tabular}[t]{l}Global\end{tabular}}}}%
    \put(0.0459436,0.11490157){\color[rgb]{0.76862745,0.50196078,0.20784314}\makebox(0,0)[lt]{\lineheight{1.25}\smash{\begin{tabular}[t]{l}Local\end{tabular}}}}%
    \put(0,0){\includegraphics[width=\unitlength,page=4]{teaser_fig_b.pdf}}%
    \put(0.79261555,0.23480946){\color[rgb]{0,0,0}\makebox(0,0)[lt]{\lineheight{1.25}\smash{\begin{tabular}[t]{l}p.\end{tabular}}}}%
    \put(0.79261555,0.11518955){\color[rgb]{0,0,0}\makebox(0,0)[lt]{\lineheight{1.25}\smash{\begin{tabular}[t]{l}p.\end{tabular}}}}%
    \put(0,0){\includegraphics[width=\unitlength,page=5]{teaser_fig_b.pdf}}%
  \end{picture}%
\endgroup%

%% file: figures/teaser_fig_c.pdf_tex
\begingroup%
  \makeatletter%
  \providecommand\color[2][]{%
    \errmessage{(Inkscape) Color is used for the text in Inkscape, but the package 'color.sty' is not loaded}%
    \renewcommand\color[2][]{}%
  }%
  \providecommand\transparent[1]{%
    \errmessage{(Inkscape) Transparency is used (non-zero) for the text in Inkscape, but the package 'transparent.sty' is not loaded}%
    \renewcommand\transparent[1]{}%
  }%
  \providecommand\rotatebox[2]{#2}%
  \newcommand*\fsize{\dimexpr\f@size pt\relax}%
  \newcommand*\lineheight[1]{\fontsize{\fsize}{#1\fsize}\selectfont}%
  \ifx\svgwidth\undefined%
    \setlength{\unitlength}{338.57238264bp}%
    \ifx\svgscale\undefined%
      \relax%
    \else%
      \setlength{\unitlength}{\unitlength * \real{\svgscale}}%
    \fi%
  \else%
    \setlength{\unitlength}{\svgwidth}%
  \fi%
  \global\let\svgwidth\undefined%
  \global\let\svgscale\undefined%
  \makeatother%
  \begin{picture}(1,0.67309487)%
    \lineheight{1}%
    \setlength\tabcolsep{0pt}%
    \put(0,0){\includegraphics[width=\unitlength,page=1]{teaser_fig_c.pdf}}%
    \put(0.37695849,0.54271999){\color[rgb]{0,0,0}\makebox(0,0)[lt]{\lineheight{1.25}\smash{\begin{tabular}[t]{l}Encoder\end{tabular}}}}%
    \put(0.00015576,0.54271999){\color[rgb]{0,0,0}\makebox(0,0)[lt]{\lineheight{1.25}\smash{\begin{tabular}[t]{l}Targets\end{tabular}}}}%
    \put(0,0){\includegraphics[width=\unitlength,page=2]{teaser_fig_c.pdf}}%
    \put(0.8903273,0.41665389){\color[rgb]{0,0,0}\makebox(0,0)[lt]{\lineheight{1.25}\smash{\begin{tabular}[t]{l}Loss\end{tabular}}}}%
    \put(0,0){\includegraphics[width=\unitlength,page=3]{teaser_fig_c.pdf}}%
    \put(0.37695849,0.17499951){\color[rgb]{0,0,0}\makebox(0,0)[lt]{\lineheight{1.25}\smash{\begin{tabular}[t]{l}Encoder\end{tabular}}}}%
    \put(0.016969,0.29433145){\color[rgb]{0.19607843,0.52941176,0.69803922}\makebox(0,0)[lt]{\lineheight{1.25}\smash{\begin{tabular}[t]{l}Global\end{tabular}}}}%
    \put(0.0459436,0.17471153){\color[rgb]{0.76862745,0.50196078,0.20784314}\makebox(0,0)[lt]{\lineheight{1.25}\smash{\begin{tabular}[t]{l}Local\end{tabular}}}}%
    \put(0.00861776,0.05509162){\color[rgb]{0.36470588,0.60784314,0.2745098}\makebox(0,0)[lt]{\lineheight{1.25}\smash{\begin{tabular}[t]{l}Cutout\end{tabular}}}}%
    \put(0,0){\includegraphics[width=\unitlength,page=4]{teaser_fig_c.pdf}}%
    \put(0.79261555,0.17721469){\color[rgb]{0,0,0}\makebox(0,0)[lt]{\lineheight{1.25}\smash{\begin{tabular}[t]{l}p.\end{tabular}}}}%
    \put(0,0){\includegraphics[width=\unitlength,page=5]{teaser_fig_c.pdf}}%
    \put(0.79261555,0.29683461){\color[rgb]{0,0,0}\makebox(0,0)[lt]{\lineheight{1.25}\smash{\begin{tabular}[t]{l}p.\end{tabular}}}}%
    \put(0,0){\includegraphics[width=\unitlength,page=6]{teaser_fig_c.pdf}}%
    \put(0.79261555,0.05759477){\color[rgb]{0,0,0}\makebox(0,0)[lt]{\lineheight{1.25}\smash{\begin{tabular}[t]{l}p.\end{tabular}}}}%
    \put(0,0){\includegraphics[width=\unitlength,page=7]{teaser_fig_c.pdf}}%
  \end{picture}%
\endgroup%

%% file: sections/related.tex
\section{Related Work}

\paragraph{SSL and Self-predictive Methods.}
Early work in self-supervised learning focused on handcrafted pretext tasks that required solving spatial or contextual prediction problems, including inpainting~\cite{pathak2016context}, jigsaw puzzles~\cite{noroozi2016unsupervisedjigsaw}, relative patch prediction~\cite{doersch2015unsupervisedcontext}, rotation prediction~\cite{gidaris2018unsupervisedrotation}, and colorization~\cite{zhang2016colorful}. These methods demonstrated that purely spatial supervision can yield semantically meaningful representations. This perspective remains relevant to our approach, which leverages an asymmetric cutout as a lightweight spatial latent-space task.

Subsequent work shifted toward invariance-based objectives. Contrastive approaches such as SimCLR~\cite{chen2020simclr} and MoCo~\cite{he2020mocov1} align augmented views while repelling different images, while clustering-based methods like DeepCluster~\cite{caron2018deep} and SwAV~\cite{caron2020swav} replace instance discrimination with prototype assignments. In parallel, non-contrastive methods explicitly enforcing feature diversity through decorrelation or variance constraints, as in Barlow Twins~\cite{zbontar2021barlow}, VICReg~\cite{bardes2021vicreg}, W-MSE~\cite{ermolov2021whitening}, and ADM~\cite{deplaen2025admin}.

Self-predictive methods \cite{grill2020byol,chen2021simsiam,chen2021mocov3} showed that strong representations can be learned using a single invariance loss, without negatives, clustering, or reconstruction. Collapse is avoided through architectural asymmetry \cite{richemond2023theedge_byol} via a predictor on the online branch. MoCo~v3~\cite{chen2021mocov3} extended this paradigm to Vision Transformers~\cite{dosovitskiy2021vit}, demonstrating the benefits of asymmetry even in contrastive formulations.
Teacher-student approaches such as DINO~\cite{caron2021dino} and iBOT~\cite{zhou2021ibot} also rely on online-target asymmetry with EMA updates, but omit a predictor head and instead stabilize training through centroid-based probability targets, or a centering and sharpening mechanism. 

While self-predictive methods are stable in the two-crop regime, naively extending them to multi-crop often leads to instability~\cite{caron2021dino,morningstar2024augsvsalgos,moon2023fMSBReg,aubret2025cobyol}. Prior work proposed addressing this issue via auxiliary regularization~\cite{zhang2022leverage}. In contrast, we show that the shared predictor itself is the source of instability: assigning one predictor per view type stabilizes multi-crop training.

Masked image modeling (MIM) constitutes another major direction, training models to reconstruct masked content either in pixel space~\cite{he2022MAE,xie2022simmim} or at the token level~\cite{bao2021beit}. Hybrid approaches such as MSN~\cite{assran2022masked} and iBOT~\cite{zhou2021ibot} further unify masked prediction with prototype-based or distillation objectives, primarily in ViT-based settings~\cite{assran2023ijepa,darcet2025CAPI}. While these works broaden the SSL design space, our focus is on improving self-predictive Siamese methods in a backbone-agnostic manner.

\paragraph{Data Augmentations in SSL.}
Data augmentations are central in Siamese SSL because they define the proxy task. Cropping is particularly powerful, as it promotes spatial correspondence across views and is responsible for most of the accuracy gains~\cite{grill2020byol,moutakanni2024noaugs,morningstar2024augsvsalgos}. 
The multi-crop strategy introduced in SwAV~\cite{caron2020swav} and adopted in DINO~\cite{caron2021dino} enriches supervision with local views, but poses challenges for self-predictive methods, which we address directly.

Several other approaches explicitly leveraged local features~\cite{xie2021detco,xiao2021resim,wang2021densecl,lebailly2023globloc}: they imposed region- or pixel-level consistency through additional losses or dense contrastive objectives. In contrast, our method follows SwAV and operates solely on final representations, introducing global-local consistency through augmentations rather than multi-level or dense supervision.
In parallel, cutout-style perturbations, including random erasing, mixup, CutMix, and object-centric cropping, have been widely used to regularize supervised and self-supervised learning~\cite{zhong2020random,zhang2017mixup,yun2019cutmix,mishra2021object}. In this work, we focus specifically on cutout as a form of within-image occlusion that isolates spatial masking effects without mixing images or labels, making it a simple building block.

Recent latent-space masked prediction frameworks such as I-JEPA~\cite{assran2023ijepa} and CAPI~\cite{darcet2025CAPI} have shown that spatial prediction can yield strong semantic representations. Our asymmetric cutout shares this spirit but differs in a key respect: rather than predicting specific masked patch representations, we predict a single global embedding, which we show is sufficient to learn useful features while remaining backbone-agnostic.

\paragraph{Multi-Task SSL.}
Combining multiple pretext tasks has long been shown to improve representation learning.
Early work~\cite{doersch2017multitaskssl} demonstrated the benefits of multi-task pretext learning with a shared backbone, and similar results were reported for 3D point clouds~\cite{hassani2019unsupervised}. Multi-task SSL has also been explored in skeleton-based action recognition with shared encoders and task-specific heads~\cite{lin2020ms2l}.
With transformer-based masked modeling, multi-task learning became closely tied to multi-modality, with methods reconstructing multiple modalities within a unified architecture~\cite{geng2022multimodal,bachmann2022multimae,yang2023comae,jamal2025multi}.
However, these methods are largely reconstruction- or modality-driven and often rely on task-specific decoders. In contrast, we operate purely in latent space and show that multi-task behavior emerges from enforcing alignment across diverse spatial view types.

Closer to our setting, adaptive multi-head contrastive learning~\cite{wang2024adaptive} and multi-target BYOL~\cite{singhbranching} introduce architectural branching for heterogeneous views but do not address predictor design or multi-crop instability. Our approach instead assigns a dedicated predictor per view type while sharing the encoder and loss, enabling stable multi-crop training and effective multi-view integration.

%% file: sections/method.tex
\section{Method}

\subsection{Background: Self-Predictive Methods}

Siamese self-supervised methods learn by aligning representations of different augmented views of the same image. Two identical encoders process the views, and a small projection head maps backbone outputs to the representation space used for the loss. Including this projector has been shown to improve the quality of representations~\cite{chen2020simclr}.

Early methods~\cite{chen2020simclr,he2020mocov1} use contrastive objectives that combine an alignment term with a repulsion term to prevent trivial solutions such as representational collapse. Later approaches, including BYOL~\cite{grill2020byol} and SimSiam~\cite{chen2021simsiam}, remove the need for explicit negative samples. Instead, they introduce asymmetry by assigning different roles to the two branches: the online branch includes a learnable predictor trained via gradient descent, while the target branch is not directly optimized and omits this predictor.
In BYOL, target parameters are updated as an exponential moving average of the online network; in SimSiam, the target is a stop-gradient copy of the online network.

\subsection{Stabilizing Multi-Crop via Decoupled Predictors} \label{sec:stabilized_multi_crop}

The multi-crop strategy extends the standard two-view setup by introducing several smaller local crops alongside the usual global views. Each crop is encoded independently, and the model learns to align local representations with their corresponding global ones. This encourages consistency across scales and contextual levels.

Multi-crop significantly improves the performance of many SSL methods, including contrastive frameworks such as SimCLR~\cite{chen2020simclr} and MoCo~\cite{he2020mocov1}, as well as clustering-based approaches like SwAV~\cite{caron2020swav} and DINO~\cite{caron2021dino}. However, this improvement is not universal. In self-predictive architectures~\cite{grill2020byol,chen2021simsiam,chen2021mocov3}, multi-crop leads to training instability and degraded performance compared to their standard baselines~\cite{caron2021dino, morningstar2024augsvsalgos, moon2023fMSBReg,aubret2025cobyol}. 
Caron \etal~\cite{caron2021dino} reported this issue for BYOL with both ResNet and Vision Transformer backbones (With a ViT-S, accuracy drops from 71.4\% to 64.8\% when using multi-crop). Morningstar \etal~\cite{morningstar2024augsvsalgos} observed analogous findings for MoCo~v3, where performance drops by 1.5\% with multi-crop, primarily because many runs exhibited training instability. While reducing the batch size or learning rate mitigated these failures, it again yielded models that underperformed the standard two-view baselines.

We hypothesize that this instability arises because global and local crops induce two fundamentally different latent-space tasks. Global-to-global alignment optimizes for augmentation invariance, whereas local-to-global alignment requires predicting global semantics from very limited contextual information (e.g., only an animal's fur). In self-predictive methods, a single prediction head is therefore required to solve both tasks simultaneously, leading to degraded performance. We refer interested readers to App. \ref{app:extended_mc_analysis} for an extended analysis. 

To resolve this issue, we assign a separate predictor to each view type while keeping the encoder shared (see~\Cref{fig:multi_task_overview_illustr}). 
Each predictor specializes in its corresponding view type, reducing interference between global and local alignment tasks. 
This modification does not alter the loss function or require new hyperparameters; it only marginally increases the number of learnable parameters and keeps training time unchanged, since each forward pass uses a single predictor. 
By isolating predictors across view types, we achieve stable optimization across different architectures.
This stabilized formulation enables self-predictive methods to benefit from multi-crop augmentations (see~\Cref{sec:results_multi_crop}) and serves as the foundation for the multi-task framework described next.

\subsection{Spatial Transformations as Latent-Space Tasks} \label{sec:multi_task_formulation}

With predictors decoupled by view type, multi-crop can be reinterpreted as a set of alignment tasks sharing a common encoder, each spatial transformation defining a distinct pre-training objective solved by its dedicated predictor. This perspective naturally suggests extending the framework by incorporating additional transformations as complementary latent-space tasks.

\begin{figure}[htb] %
    \centering
    \def\svgwidth{0.41\columnwidth}
    {\scriptsize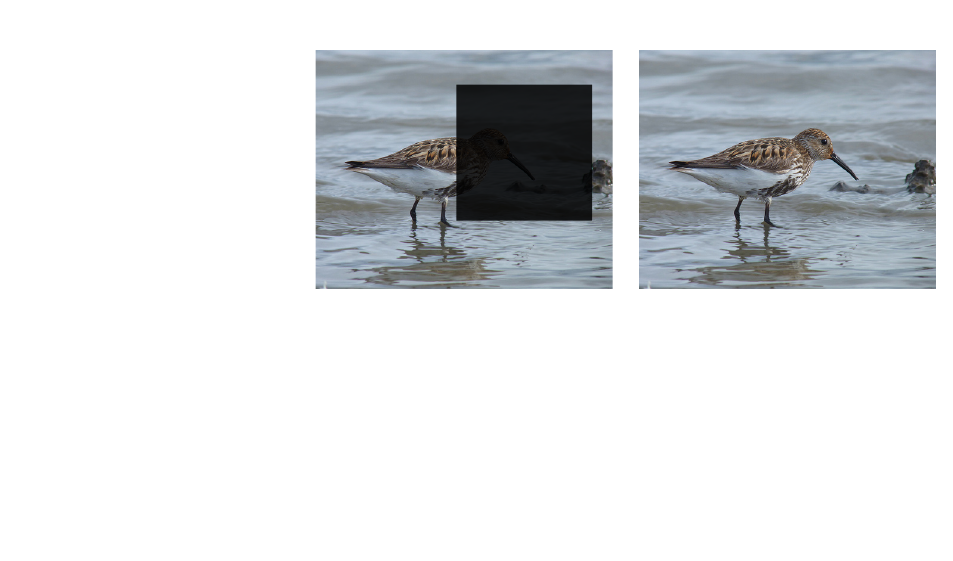}
    \caption{Asymmetric and symmetric cutout. Image from ImageNet val. set (\textnumero 7011).}
    \label{fig:asym_cutout_example}
\end{figure}

We introduce asymmetric cutout~\cite{devries2017cutout} as one such task. While local crops require the model to infer global semantics from a reduced spatial context, random cutouts mask internal regions, encouraging robustness to partial occlusions. 
We apply cutout asymmetrically (see~\Cref{fig:asym_cutout_example}): the online branch receives the masked view while the target remains complete. In this setup, the online model is trained to predict the representation of the full image from a partially masked view, forming a semantic inpainting objective in latent space. 

Formally, let $\vz^{v}$ denote the representation of view type $v\in \{\text{glob},\text{loc},\text{cutout}\}$. The total alignment loss is computed as a weighted sum over view types:
\begin{equation}
\mathcal{L}=\sum_v \lambda_v \mathbb{E}\left[\left\|q_v\left(\vz_v\right)-\vz_{\text {glob}}\right\|_2^2\right]
\end{equation}
where $q_v$ denotes the predictor for view type $v$ and $\lambda_v$ its relative weight. 
To keep the method simple and avoid tuning for dataset-specific statistics, we simply set all task weights to the same value: $\lambda_{\text{glob}} = \lambda_{\text{loc}} = \lambda_{\text{cutout}}$. 

\begin{figure}[hbt] %
    \centering
    \def\svgwidth{0.96\textwidth}
    {\scriptsize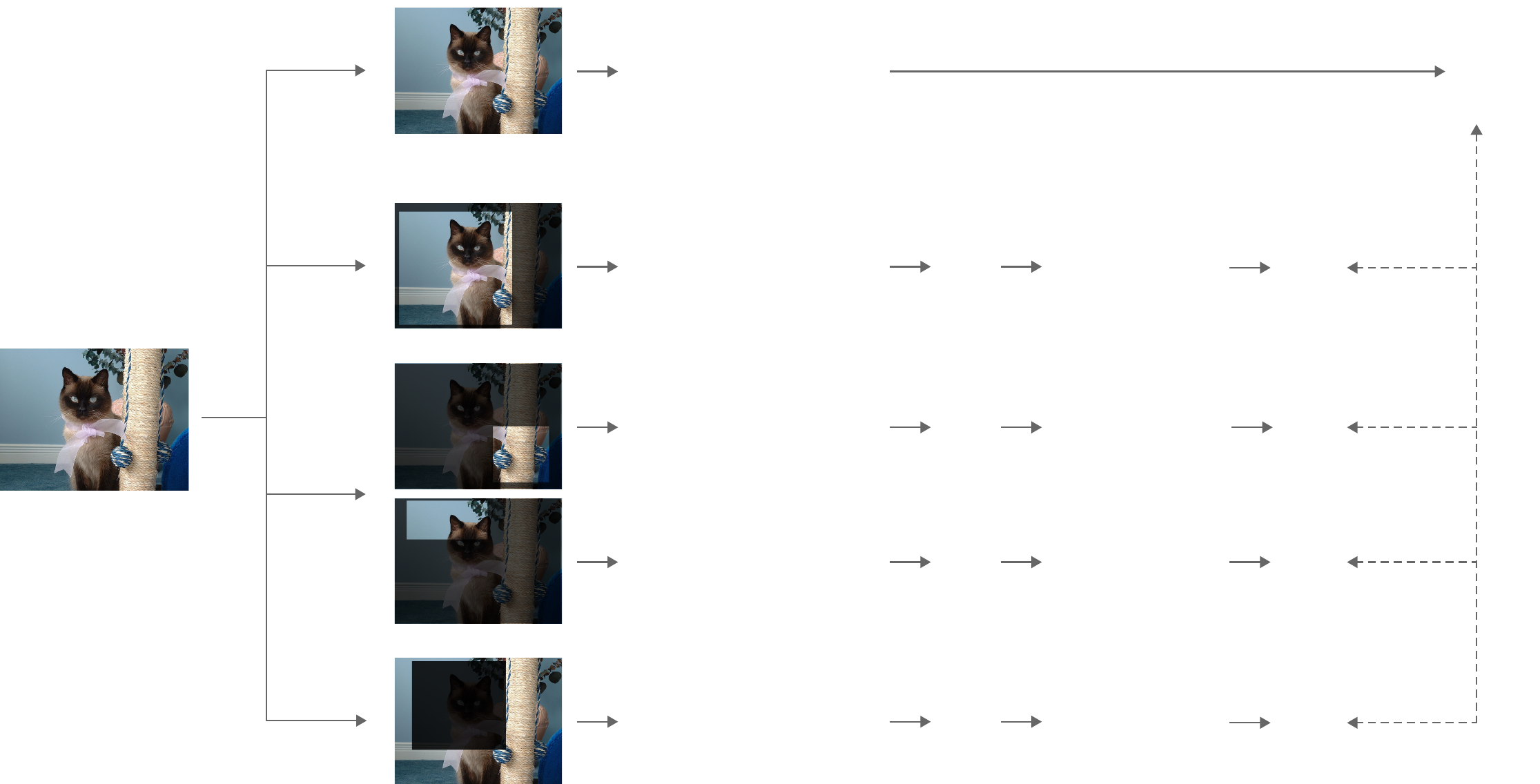}
    \caption{The proposed MULAN framework. A shared encoder processes all spatial views (global, local, cutout), while task-specific prediction heads align each view's representation with the target. All tasks are optimized jointly under the same loss. Image from ImageNet val. set (\textnumero 43632).}
    \label{fig:multi_task_overview_illustr}
\end{figure}

This unified formulation (\Cref{fig:multi_task_overview_illustr}) treats each spatial transformation as a distinct latent-space task with its own predictor, while sharing the encoder and alignment loss across all view types. It integrates global, local, and cutout views into a single stable training recipe, is compatible across backbone architectures, and provides a principled way to incorporate additional spatial tasks in the future. We refer to this unified framework as MULAN (multi-task latent-space network).

%% file: figures/cutout_sym_vs_asym_v2.pdf_tex
\begingroup%
  \makeatletter%
  \providecommand\color[2][]{%
    \errmessage{(Inkscape) Color is used for the text in Inkscape, but the package 'color.sty' is not loaded}%
    \renewcommand\color[2][]{}%
  }%
  \providecommand\transparent[1]{%
    \errmessage{(Inkscape) Transparency is used (non-zero) for the text in Inkscape, but the package 'transparent.sty' is not loaded}%
    \renewcommand\transparent[1]{}%
  }%
  \providecommand\rotatebox[2]{#2}%
  \newcommand*\fsize{\dimexpr\f@size pt\relax}%
  \newcommand*\lineheight[1]{\fontsize{\fsize}{#1\fsize}\selectfont}%
  \ifx\svgwidth\undefined%
    \setlength{\unitlength}{457.12561526bp}%
    \ifx\svgscale\undefined%
      \relax%
    \else%
      \setlength{\unitlength}{\unitlength * \real{\svgscale}}%
    \fi%
  \else%
    \setlength{\unitlength}{\svgwidth}%
  \fi%
  \global\let\svgwidth\undefined%
  \global\let\svgscale\undefined%
  \makeatother%
  \begin{picture}(1,0.58837467)%
    \lineheight{1}%
    \setlength\tabcolsep{0pt}%
    \put(0,0){\includegraphics[width=\unitlength,page=1]{cutout_sym_vs_asym_v2.pdf}}%
    \put(0.39068103,0.56095084){\makebox(0,0)[lt]{\lineheight{1.25}\smash{\begin{tabular}[t]{l}Online view\end{tabular}}}}%
    \put(0.72964694,0.56095084){\makebox(0,0)[lt]{\lineheight{1.25}\smash{\begin{tabular}[t]{l}Target view\end{tabular}}}}%
    \put(0.02804001,0.37145151){\color[rgb]{0,0.52941176,0.09803922}\makebox(0,0)[lt]{\lineheight{1.25}\smash{\begin{tabular}[t]{l}(inpainting-like)\end{tabular}}}}%
    \put(0.0411491,0.42067212){\color[rgb]{0,0.52941176,0.09803922}\makebox(0,0)[lt]{\lineheight{1.25}\smash{\begin{tabular}[t]{l}\footnotesize Asymmetric\end{tabular}}}}%
    \put(0,0){\includegraphics[width=\unitlength,page=2]{cutout_sym_vs_asym_v2.pdf}}%
    \put(0.05671922,0.08597199){\color[rgb]{0.68627451,0,0}\makebox(0,0)[lt]{\lineheight{1.25}\smash{\begin{tabular}[t]{l}(invariance)\end{tabular}}}}%
    \put(0.04889314,0.1351926){\color[rgb]{0.68627451,0,0}\makebox(0,0)[lt]{\lineheight{1.25}\smash{\begin{tabular}[t]{l}\footnotesize Symmetric\end{tabular}}}}%
  \end{picture}%
\endgroup%

%% file: figures/MULAN_overview.pdf_tex
\begingroup%
  \makeatletter%
  \providecommand\color[2][]{%
    \errmessage{(Inkscape) Color is used for the text in Inkscape, but the package 'color.sty' is not loaded}%
    \renewcommand\color[2][]{}%
  }%
  \providecommand\transparent[1]{%
    \errmessage{(Inkscape) Transparency is used (non-zero) for the text in Inkscape, but the package 'transparent.sty' is not loaded}%
    \renewcommand\transparent[1]{}%
  }%
  \providecommand\rotatebox[2]{#2}%
  \newcommand*\fsize{\dimexpr\f@size pt\relax}%
  \newcommand*\lineheight[1]{\fontsize{\fsize}{#1\fsize}\selectfont}%
  \ifx\svgwidth\undefined%
    \setlength{\unitlength}{1070.86640625bp}%
    \ifx\svgscale\undefined%
      \relax%
    \else%
      \setlength{\unitlength}{\unitlength * \real{\svgscale}}%
    \fi%
  \else%
    \setlength{\unitlength}{\svgwidth}%
  \fi%
  \global\let\svgwidth\undefined%
  \global\let\svgscale\undefined%
  \makeatother%
  \begin{picture}(1,0.50883664)%
    \lineheight{1}%
    \setlength\tabcolsep{0pt}%
    \put(0.04006102,0.29182912){\color[rgb]{0,0,0}\makebox(0,0)[lt]{\lineheight{1.25}\smash{\begin{tabular}[t]{l}Image\end{tabular}}}}%
    \put(0,0){\includegraphics[width=\unitlength,page=1]{MULAN_overview.pdf}}%
    \put(0.18136716,0.4717045){\color[rgb]{0,0,0}\makebox(0,0)[lt]{\lineheight{1.25}\smash{\begin{tabular}[t]{l}Targets\end{tabular}}}}%
    \put(0.18136716,0.34544225){\color[rgb]{0.19607843,0.52941176,0.69803922}\makebox(0,0)[lt]{\lineheight{1.25}\smash{\begin{tabular}[t]{l}Task A\end{tabular}}}}%
    \put(0.18136716,0.19696434){\color[rgb]{0.76862745,0.50196078,0.20784314}\makebox(0,0)[lt]{\lineheight{1.25}\smash{\begin{tabular}[t]{l}Task B\end{tabular}}}}%
    \put(0.18136716,0.04918681){\color[rgb]{0.36470588,0.60784314,0.2745098}\makebox(0,0)[lt]{\lineheight{1.25}\smash{\begin{tabular}[t]{l}Task C\end{tabular}}}}%
    \put(0,0){\includegraphics[width=\unitlength,page=2]{MULAN_overview.pdf}}%
    \put(0.70147405,0.33969924){\color[rgb]{0,0,0}\makebox(0,0)[lt]{\lineheight{1.25}\smash{\begin{tabular}[t]{l}Prediction\end{tabular}}}}%
    \put(0,0){\includegraphics[width=\unitlength,page=3]{MULAN_overview.pdf}}%
    \put(0.97098947,0.22635176){\color[rgb]{0,0,0}\makebox(0,0)[lt]{\lineheight{1.25}\smash{\begin{tabular}[t]{l}loss\end{tabular}}}}%
    \put(0.71215466,0.31950064){\color[rgb]{0,0,0}\makebox(0,0)[lt]{\lineheight{1.25}\smash{\begin{tabular}[t]{l}Head A\end{tabular}}}}%
    \put(0.70147405,0.23534448){\color[rgb]{0,0,0}\makebox(0,0)[lt]{\lineheight{1.25}\smash{\begin{tabular}[t]{l}Prediction\end{tabular}}}}%
    \put(0.71274297,0.21514588){\color[rgb]{0,0,0}\makebox(0,0)[lt]{\lineheight{1.25}\smash{\begin{tabular}[t]{l}Head B\end{tabular}}}}%
    \put(0.70147405,0.14919928){\color[rgb]{0,0,0}\makebox(0,0)[lt]{\lineheight{1.25}\smash{\begin{tabular}[t]{l}Prediction\end{tabular}}}}%
    \put(0.71274297,0.12900069){\color[rgb]{0,0,0}\makebox(0,0)[lt]{\lineheight{1.25}\smash{\begin{tabular}[t]{l}Head B\end{tabular}}}}%
    \put(0.70147405,0.0455449){\color[rgb]{0,0,0}\makebox(0,0)[lt]{\lineheight{1.25}\smash{\begin{tabular}[t]{l}Prediction\end{tabular}}}}%
    \put(0.7124208,0.0253463){\color[rgb]{0,0,0}\makebox(0,0)[lt]{\lineheight{1.25}\smash{\begin{tabular}[t]{l}Head C\end{tabular}}}}%
    \put(0.447892,0.46736922){\color[rgb]{0,0,0}\makebox(0,0)[lt]{\lineheight{1.25}\smash{\begin{tabular}[t]{l}Backbone +\end{tabular}}}}%
    \put(0.43342941,0.44713561){\color[rgb]{0,0,0}\makebox(0,0)[lt]{\lineheight{1.25}\smash{\begin{tabular}[t]{l}Projection Head\end{tabular}}}}%
    \put(0.447892,0.33990234){\color[rgb]{0,0,0}\makebox(0,0)[lt]{\lineheight{1.25}\smash{\begin{tabular}[t]{l}Backbone +\end{tabular}}}}%
    \put(0.43342941,0.31966873){\color[rgb]{0,0,0}\makebox(0,0)[lt]{\lineheight{1.25}\smash{\begin{tabular}[t]{l}Projection Head\end{tabular}}}}%
    \put(0.447892,0.23554759){\color[rgb]{0,0,0}\makebox(0,0)[lt]{\lineheight{1.25}\smash{\begin{tabular}[t]{l}Backbone +\end{tabular}}}}%
    \put(0.43342941,0.21531397){\color[rgb]{0,0,0}\makebox(0,0)[lt]{\lineheight{1.25}\smash{\begin{tabular}[t]{l}Projection Head\end{tabular}}}}%
    \put(0.447892,0.14870202){\color[rgb]{0,0,0}\makebox(0,0)[lt]{\lineheight{1.25}\smash{\begin{tabular}[t]{l}Backbone +\end{tabular}}}}%
    \put(0.43342941,0.12846841){\color[rgb]{0,0,0}\makebox(0,0)[lt]{\lineheight{1.25}\smash{\begin{tabular}[t]{l}Projection Head\end{tabular}}}}%
    \put(0.447892,0.045748){\color[rgb]{0,0,0}\makebox(0,0)[lt]{\lineheight{1.25}\smash{\begin{tabular}[t]{l}Backbone +\end{tabular}}}}%
    \put(0.43342941,0.02551439){\color[rgb]{0,0,0}\makebox(0,0)[lt]{\lineheight{1.25}\smash{\begin{tabular}[t]{l}Projection Head\end{tabular}}}}%
    \put(0.93518668,0.49760409){\color[rgb]{0,0,0}\makebox(0,0)[lt]{\lineheight{1.25}\smash{\begin{tabular}[t]{l}Repres.\end{tabular}}}}%
    \put(0.8220213,0.47869416){\color[rgb]{0,0,0}\makebox(0,0)[lt]{\lineheight{1.25}\smash{\begin{tabular}[t]{l}stop-grad\end{tabular}}}}%
    \put(0.60398991,0.37143289){\color[rgb]{0,0,0}\makebox(0,0)[lt]{\lineheight{1.25}\smash{\begin{tabular}[t]{l}Repres.\end{tabular}}}}%
    \put(0.50680692,0.39384465){\color[rgb]{0,0,0}\makebox(0,0)[lt]{\lineheight{1.25}\smash{\begin{tabular}[t]{l}EMA / copy\end{tabular}}}}%
    \put(0.81707671,0.37083757){\color[rgb]{0,0,0}\makebox(0,0)[lt]{\lineheight{1.25}\smash{\begin{tabular}[t]{l}Prediction\end{tabular}}}}%
  \end{picture}%
\endgroup%

%% file: sections/experiments.tex
\section{Experiments}

\subsection{Validation of Multi-Predictor Multi-Crop} \label{sec:results_multi_crop}

We first validate our solution to the instability of multi-crop training in self-predictive frameworks. 
As discussed in \Cref{sec:stabilized_multi_crop}, a shared predictor must align representations from both global and local crops, often causing instability. 
To address this, we assign a separate predictor to each view type while keeping the encoder shared, and train BYOL, SimSiam, and MoCo~v3 under identical settings.

\paragraph{Setup.}
All experiments use a ResNet-50~\cite{he2016resnet} backbone and follow the original training recipes of each method, except that batch sizes are reduced to 1024 for BYOL and MoCo v3\footnote{\label{footnote:batchsize}This allows training on a single 4-GPU node, with only marginal impact on accuracy.}. Models are trained for 200 epochs on ImageNet-1k~\cite{deng2009imagenet} with standard augmentations. 
Representation quality is measured using (1) linear evaluation, where a linear classifier is trained on frozen backbone features, and (2) non-parametric $k$-nearest neighbor (kNN) evaluation on the embedding space.

\paragraph{Results.}
\Cref{tab:multi-crop_rocks} reports linear and kNN accuracies. Introducing view-specific predictors lets all three frameworks benefit from multi-crop training. In BYOL, linear accuracy rises from 70.7\% to 74.7\%. SimSiam and MoCo~v3 show similar gains of around +4 points each. All runs converge reliably at standard learning rates, confirming that the shared predictor was the main source of instability.
These results demonstrate that self-predictive SSL methods can fully benefit from multi-crop augmentations once predictors are decoupled by view type. The modification is lightweight and generalizes across frameworks. 

\begin{table}[htb]
\small
\caption{Downstream performance of self-supervised ResNet-50 models trained with the multi-predictor multi-crop (m-c) formulation. All models are pre-trained for 200 epochs.}
\label{tab:multi-crop_rocks}
\centering
\begin{tabular}{lcccc}
    \toprule
    Method & kNN & $\Delta$ & lin. & $\Delta$ \\ \midrule
    BYOL & 63.1 & & 70.7 & \\ 
    \quad w. multi-predictor m-c & \textbf{69.5} & {\color{greentxt}+6.4} & \textbf{74.7} & {\color{greentxt}+4.0} \\ %
    \midrule
    SimSiam & 60.2 & & 69.9 &  \\
    \quad w. multi-predictor m-c & \textbf{65.3} & {\color{greentxt}+4.9} & \textbf{73.9} & {\color{greentxt}+4.0}  \\ %
    \midrule
    MoCo v3 & 64.5 & & 71.4 &  \\
    \quad w. multi-predictor m-c & \textbf{68.7} & {\color{greentxt}+4.2} & \textbf{75.2} & {\color{greentxt}+3.8} \\ %
    \bottomrule
\end{tabular}
\end{table}

\subsection{Asymmetric Cutout as a Standalone Task}
\label{sec:assym_cutout_validation}

Before adding cutout views to our multi-task formulation, we examine whether random cutout alone provides a valuable learning signal. 
We train BYOL using cutout as the only augmentation, masking a random region in the online view while keeping the target view complete. 
After 100 epochs on ImageNet, this simple setup reaches 46.8\% accuracy, confirming that the model can learn meaningful representations by predicting the embedding of a full image from a partially masked view.

We then compare asymmetric and symmetric masking. When both views are masked, the model achieves near-random performance (3.4\%). 
This confirms that asymmetry is crucial: the model must infer missing information from an unmasked reference rather than from another incomplete view. 
These results validate asymmetric cutout as a viable latent-space task and motivate its integration in our final multi-task formulation.

\subsection{Evaluation of the Multi-Task Formulation} \label{sec:results_multi-task}

We now evaluate whether the proposed asymmetric cutout latent-space objective (\Cref{sec:multi_task_formulation}) complements the two existing latent-space tasks (\Cref{sec:stabilized_multi_crop}). Specifically, we assess whether models can leverage this additional pre-training signal to improve representation quality. We apply the multi-task formulation to BYOL, SimSiam, and MoCo v3, and compare each method against its multi-predictor multi-crop baseline under matched computational budgets.

\paragraph{Implementation details.}  
We pre-train ResNet-50 backbones following the official training recipes of each method (see App.~\ref{app:expe_setups} for full settings). For a fair comparison, we adjust the number and type of views so that the total computational cost remains similar; specifically, we replace two local views with a single cutout view, since local views require roughly half the computation. 

\paragraph{Results.}  
\Cref{tab:full_comparison} shows that adding cutout views further improves the performance of the three multi-crop frameworks by a significant margin. 
Even with only 200 pre-training epochs, all three multi-task models outperform their respective baselines. 
For instance, the multi-task SimSiam variant reaches 74.7\% linear accuracy after only 200 pre-training epochs, surpassing the 71.3\% achieved by the original SimSiam even after 800 epochs.
These gains also translate into improved training efficiency. In the case of BYOL, the multi-task variant reaches 75.6\% linear accuracy after 200 epochs (72 hours), surpassing the 74.3\% achieved by the 1000-epoch baseline despite requiring only about one-third of the total training time (219h), even though its per-epoch cost is slightly higher (21:35 vs. 13:09) (see App.~\ref{app:timing}, \Cref{tab:efficiency}).

\begin{table}[htb]
\small
\centering
\caption{Downstream performance of self-supervised ResNet-50 models under different augmentation strategies. Gray baselines correspond to the full-schedule results from the original papers.}
\label{tab:full_comparison}
\begin{tabular}{lllcc}
\toprule
Method & Strategy & Epochs & kNN & lin. \\ \midrule
{\color{gray} BYOL} & {\color{gray} baseline} & {\color{gray} 1000} & {\color{gray} 68.0} & {\color{gray} 74.3} \\
BYOL & baseline & 200 & 63.1 & 70.7 \\
 & multi-predictor m-c & 200 & 69.5 &   74.7   \\ 
 & multi-task & 200 & \textbf{69.9}  & \textbf{75.6} \\ %
\midrule
{\color{gray} SimSiam} & {\color{gray} baseline} & {\color{gray} 800} & {\color{gray} --} & {\color{gray} 71.3} \\
SimSiam & baseline & 200 & 60.2 & 69.9 \\
 & multi-predictor m-c & 200 & 65.3 & 73.9  \\
 & multi-task & 200 & \textbf{66.5} & \textbf{74.7} \\ %
\midrule
{\color{gray} MoCo v3} & {\color{gray} baseline} & {\color{gray} 1000} & {\color{gray} 68.9} & {\color{gray} 74.6} \\
MoCo v3 & baseline & 200 & 64.5 & 71.4 \\
 & multi-predictor m-c & 200 & 68.7 & 75.2 \\ 
 & multi-task & 200 & \textbf{69.2} & \textbf{75.7}  \\ %
\bottomrule
\end{tabular}
\end{table}

\subsection{Comparison with State-of-the-Art Methods} \label{sec:sota_results}

To assess the scalability to longer training schedules, competitiveness, and general applicability of our approach across backbones, we train MULAN on ImageNet-1k~\cite{deng2009imagenet} using both convolutional and transformer backbones. 

\paragraph{Implementation details.} 
We adopt BYOL~\cite{grill2020byol} as the base framework, given its strong results in our 200-epoch experiments. For the \textbf{ResNet-50}~\cite{he2016resnet} backbone, we follow the official BYOL configuration: two-layer projector and predictor networks (hidden dimension 4096, output 256), each with intermediate batch normalization~\cite{ioffe2015batchnorm} and ReLU activations. We pre-train the model for 800 epochs with the LARS optimizer~\cite{you2017LARS}, a linearly scaled~\cite{goyal2017accurate} learning rate ($lr = 0.4 \times \text{batch\_size} / 256$), a cosine decay schedule, and a 10-epoch warm-up. The batch size is 1024\footref{footnote:batchsize}, and the weight decay is $1.5\times10^{-6}$, excluding batch normalization and bias parameters. The target network is updated using an EMA of the student weights~\cite{lillicrap2015continuous}, with the momentum coefficient increasing from 0.996 to 1.0 following a cosine schedule.
For the \textbf{ViT}~\cite{dosovitskiy2021vit} backbones, we mostly follow the MoCo~v3~\cite{chen2021mocov3} recipe and head design, except that we do not freeze the patch projection layer, as it led to lower performance in the multi-task setting. Instead, we found that lowering Adam's $\beta_2$ together with gradient clipping mitigated training instability (see App. \ref{app:vit_details}). We pre-train the backbone for 200 epochs using the AdamW optimizer~\cite{kingma2014adam}, a linearly scaled learning rate (base $3\times10^{-4}$), a batch size of 1024, and a weight decay of 0.1. The EMA base value is set to 0.998. 

\begin{table}[!htb]
\small
\centering
\caption{Evaluation of SSL techniques pre-trained on ImageNet-1k. $^\dag$MIM methods. $^\ddag$hybrid approaches. $^\star$Checkpoints converted from JAX.}
\label{tab:results_imagenet_ssl}
\begin{minipage}{0.48\textwidth}
\centering
\begin{tabular}{rlll} 
\toprule
    Ref. & Method & kNN & Lin. \\
\midrule
    \multicolumn{4}{c}{\textit{Backbone: ResNet-50}} \\
    \cite{vryniotis2021pytorch_resnet} & {\color{gray} supervised} & \multicolumn{1}{c}{\color{gray} --} & {\color{gray} 80.5} \\
    \cite{chen2020simclr} & SimCLR & 60.7 & 69.3 \\
    \cite{chen2020mocov2} & MoCo v2 & 61.9 & 71.1 \\
    \cite{chen2021simsiam} & SimSiam & \multicolumn{1}{c}{--} & 71.3 \\
    \cite{zbontar2021barlow} & Barlow Twins & 66.0 & 73.2 \\
    \cite{grill2020byol} & BYOL & 68.0$^\star$ & 74.3 \\
    \cite{chen2021mocov3} & MoCo v3 & 68.9 & 74.6  \\
    \cite{caron2020swav} & SwAV & 65.7 & 75.3  \\
    \cite{caron2021dino} & DINO & 67.5 & 75.3  \\
    \cite{lee2021cbyol} & C-BYOL & \multicolumn{1}{c}{--} & 75.6 \\
    \cite{tomasev2022relicv2} & ReLIC v2 & 70.5$^\star$ & \textbf{77.1} \\
    \rowcolor{lightblue}
    & MULAN & \textbf{70.9} & 76.7 \\
\bottomrule
\end{tabular}
\end{minipage}
\hfill
\begin{minipage}{0.48\textwidth}
\centering
\begin{tabular}{rlll} 
\toprule
    Ref. & Method & kNN & Lin. \\
\midrule
    \multicolumn{4}{c}{\textit{Backbone: ViT-S}} \\
    \cite{touvron2021deit} & {\color{gray} supervised} & \multicolumn{1}{c}{\color{gray} --} & {\color{gray} 79.8} \\
    \cite{grill2020byol} & BYOL & 66.6 & 71.4 \\
    \cite{chen2020mocov2} & MoCo v2 & 64.4 & 72.7 \\ 
    \cite{chen2021mocov3} & MoCo v3 & \multicolumn{1}{c}{--} & 73.4 \\
    \cite{caron2020swav} & SwAV & 66.3 & 73.5 \\ 
    \cite{caron2021dino} & DINO & \textbf{74.5} & \textbf{77.0} \\
    \rowcolor{lightblue} & MULAN & 70.2 & 74.5 \\
\midrule
    \multicolumn{4}{c}{\textit{Backbone: ViT-B}} \\ 
    \cite{touvron2021deit} & {\color{gray} supervised} & \multicolumn{1}{c}{\color{gray} --} & {\color{gray} 81.8} \\
    \cite{xie2022simmim} & SimMIM$^\dag$   & 16.1 & 56.7 \\
    \cite{he2022MAE} & MAE$^\dag$      & 27.1 & 68.0 \\
    \cite{assran2023ijepa} & I-JEPA$^\dag$   & \multicolumn{1}{c}{--} & 72.9 \\
    \cite{zhou2021ibot} & iBOT$^\ddag$ & 77.1 & 79.5 \\
    \cdashline{1-4}[2pt/4pt] \addlinespace[2pt]
    \cite{chen2020simclr} & SimCLR   & \multicolumn{1}{c}{--} & 73.9 \\
    \cite{grill2020byol} & BYOL     & 68.1 & 73.9 \\
    \cite{chen2021mocov3} & MoCo v3  & 71.4 & 76.7 \\
    \cite{caron2021dino} & DINO      & \textbf{76.1} & 78.2 \\
    \rowcolor{lightblue} & MULAN  & 74.2 & \textbf{78.3} \\
\bottomrule
\end{tabular}
\end{minipage}
\end{table}

\paragraph{ImageNet results.} \Cref{tab:results_imagenet_ssl} summarizes the results. Our MULAN framework significantly improves upon the BYOL baseline across both CNN and transformer backbones.
\newline
With a \textbf{ResNet-50}, it achieves 76.7\% linear and 70.9\% kNN accuracy, outperforming BYOL by +2.4\% and +2.9\%, respectively, and surpassing clustering-based methods such as SwAV and DINO. Performance is comparable to ReLIC~v2, which attains slightly higher linear accuracy (77.1\%) but a lower kNN score (70.5\%). 
On \textbf{ViT} backbones, MULAN also yields consistent gains. With ViT-S, linear accuracy improves from 71.4\% to 74.5\% (+3.1\%), and scaling to ViT-B further raises it to 78.3\%, matching DINO. Notably, our method gains +3.8\% from ViT-S to ViT-B compared to only +1.2\% for DINO, suggesting MULAN interacts favorably with larger model capacity. Unlike DINO, which benefits from 800-epoch schedules, our method does not improve beyond 200 epochs on ViT backbones, yet already matches DINO's accuracy. Thus, investigating how to unlock longer-schedule training for self-predictive transformer methods is a promising direction for future work.
Finally, our approach outperforms MIM-based methods on kNN and linear evaluation, consistent with evidence that MIM representations require full fine-tuning to reach peak performance~\cite{he2022MAE,marks2025closer}. 
Overall, MULAN demonstrates consistent scalability across CNN and transformer backbones of varying capacities, yielding significant improvements in representation quality.

\paragraph{Transfer to dense tasks.}  
We evaluate transfer to COCO object detection and segmentation (\Cref{tab:transfer_coco_resnet50}) using Mask R-CNN with FPN and a ResNet-50 backbone. Dense tasks require spatially detailed features, providing a complementary assessment of learned representations beyond classification. 
The MULAN framework outperforms supervised pretraining (41.8 vs.\ 39.0 AP for detection, 38.0 vs.\ 35.4 AP for segmentation) and prior SSL methods, achieving the highest AP on both tasks. 
These results suggest that the multi-task formulation produces representations that generalize well and could be used in a variety of downstream tasks. 

We further report semi-supervised learning and fine-tuning results in App.~\ref{app:additional_evals}.

\begin{table}[htb]
\small
\centering
\begin{minipage}[t]{0.42\linewidth}
\centering
\caption{Transfer performance (ResNet-50) on COCO using Mask R-CNN with FPN (1$\times$ schedule from detectron2).}
\label{tab:transfer_coco_resnet50}
\begin{tabular}{lrr}
    \toprule
    Method & AP\textsuperscript{det.} & AP\textsuperscript{segm.} \\
    \midrule
    {\color{gray} supervised} & {\color{gray} 39.0} & {\color{gray} 35.4} \\
    MoCo v2 & 39.8 & 36.1 \\
    BYOL & 40.4 & 37.0 \\
    SwAV & 41.6 & 37.8 \\
    Barlow Twins & 40.0 & 36.7 \\
    DINO & 41.2 & 37.1 \\
    \rowcolor{lightblue} MULAN & \textbf{41.8} & \textbf{38.0} \\
    \bottomrule
\end{tabular}
\end{minipage}
\hfill
\begin{minipage}[t]{0.55\linewidth}
\centering
\caption{Effect of number and type of views on downstream performance. Each view type uses a different prediction head. Models are pre-trained for 200 epochs.}
\label{tab:ablation_views}
\begin{tabular}{lrrrrr}
    \toprule
     & glob. & loc. & cutout & kNN & Lin. \\
    \midrule
    BYOL & 2 & 0 & 0 & 63.1 & 70.7 \\
    \quad w. more views & 4 & 0 & 0 & 64.3 & 71.7 \\
    \quad w. local views & 2 & 4 & 0 & 69.5 & 74.7 \\
    \quad w. cutout views & 2 & 0 & 2 & 68.6 & 73.7 \\
    \quad w. multi-task & 2 & 2 & 1 & \textbf{69.9} & \textbf{75.6} \\
    \bottomrule
\end{tabular}
\end{minipage}
\end{table}

\begin{table}[htb]
\small
\centering
\caption{Effect of individual augmentations on BYOL (ResNet-50). Removing cropping causes a large performance drop, while cropping or asymmetric cutout alone remains competitive. Spatial augmentations outperform all non-spatial combinations. Models are trained for 200 epochs, except the cutout-only variant, which saturates earlier.}
\label{tab:augs_analysis}
\begin{tabular}{lcccccccr}
\toprule
 & \multicolumn{3}{c}{Spatial augs.} & \multicolumn{4}{c}{Other augs.} & \multirow{2}{*}{Lin acc.} \\
 \cmidrule(lr){2-4} \cmidrule(lr){5-8}
 & crop & cutout & flip & jitter & gray & solar & blur \\ \midrule
Baseline & \cmark & \xmark & \cmark & \cmark & \cmark & \cmark & \cmark & 70.7 \\ \midrule
Remove crop & \xmark & \xmark & \cmark & \cmark & \cmark & \cmark & \cmark & 33.8 \\
Crop only & \cmark & \xmark & \xmark & \xmark & \xmark & \xmark & \xmark & 55.3 \\
Cutout only & \xmark & \cmark & \xmark & \xmark & \xmark & \xmark & \xmark & 46.8 \\
\bottomrule
\end{tabular}
\end{table}

\section{Ablation Studies}

\subsection{Influence of View Composition} \label{sec:ablation_studies}

\Cref{tab:ablation_views} analyzes how the number and type of views affect BYOL performance. Increasing the number of global views from two to four yields only a modest improvement (70.7\% to 71.7\%), an effect that would likely diminish with longer training schedules. Introducing new view types instead produces substantial gains: local views raise performance to 74.7\%, and global+cutout views reach 73.7\%, confirming that view diversity is more valuable than quantity alone.
Notably, the global+cutout configuration significantly outperforms the baseline, suggesting that the benefits of multi-crop strategies can be attributed to the increased spatial diversity, rather than to the inclusion of smaller-resolution crops specifically.
The multi-task configuration achieves the best result with 75.6\% linear accuracy despite using fewer total views than the multi-crop setup. This confirms that cutout and local views provide complementary training signals. 

\subsection{On the Importance of Spatial Augmentations} \label{sec:spatial_augs_study}

We introduced asymmetric cutout as a complementary task within our multi-task formulation, motivated by the disproportionate importance of spatial augmentations in SSL. While early work by Chen \etal~\cite{chen2020simclr} emphasized that strong augmentation compositions are crucial to define invariances, subsequent studies showed that cropping alone remains remarkably competitive. For instance, BYOL achieves 59.4\% accuracy with only cropping, a trend also reported by Moutakanni \etal~\cite{moutakanni2024noaugs} for DINOv2 at scale.

Ablation results in \Cref{tab:augs_analysis} confirm that spatial transformations provide the primary training signal for SSL. In a 200-epoch schedule, cropping alone yields 55.3\% accuracy on ImageNet, while removing cropping from the full augmentation pipeline reduces accuracy from 70.7\% to 33.8\%. Additionally, our asymmetric cutout strategy, which masks only the online view, achieves 46.8\% accuracy and outperforms the combination of all non-spatial augmentations. 

However, the modularity of our framework does not imply that all spatial augmentations can serve as effective latent-space tasks. 
As shown in App. \ref{app:negative_results}, simple transformations like random rotation yielded negligible additional improvements. We hypothesize that a latent-space task must be both semantically meaningful and sufficiently challenging to push the model to extract non-trivial features. We find that validating new tasks in a standalone setting (\Cref{sec:assym_cutout_validation}) is a reliable prerequisite to successful multi-task integration.

%% file: sections/conclusion.tex
\section{Conclusion}

The key insight of this work is simple: spatial transformations in SSL are not just augmentations but latent-space tasks, and treating them as such unlocks substantial gains. Decoupling predictors by view type resolves the long-standing instability of multi-crop in self-predictive methods (BYOL, SimSiam, and MoCo v3), without changing the loss, backbone, or hyperparameters. Reframing this design as a multi-task objective then suggests incorporating additional pretext tasks. By adding \textit{asymmetric cutout} as a complementary semantic inpainting task, we further improve accuracy across architectures.
Beyond the specific gains reported, we believe the multi-task perspective offers a principled lens for future augmentation design in SSL. A natural next step is conditioning predictors on view-specific metadata such as cutout coordinates or crop scale, moving toward richer, task-conditional self-supervision. Extending this framework to video or 3D point clouds, where spatial transformations are even more diverse, is another promising direction.

%% file: sections/acknowledgments.tex
\section*{Acknowledgments}
Most of the computational resources and services used in this work were provided by the VSC (Flemish Supercomputer Center), funded by the Research Foundation Flanders (FWO) and the Flemish Government – department WEWIS.

%% file: sections/suppl.tex
\section{Experimental Settings} \label{app:expe_setups}

In this section, we detail the implementation settings required to reproduce the results reported in \Cref{sec:results_multi_crop,sec:assym_cutout_validation,sec:results_multi-task,sec:sota_results,sec:ablation_studies,sec:spatial_augs_study}.
Unless specified otherwise, all experiments use ResNet-50 backbones, the standard ImageNet-1k training/validation splits, and synchronized batch normalization. All models are trained from scratch without labels, using mixed-precision training on a single 4$\times$Nvidia A100 GPU node.

\paragraph{BYOL and MoCo v3 baselines.} We closely follow the official training recipes for both methods, which use nearly identical hyperparameters. 
We pre-train ResNet-50 backbones with two-layer projection and prediction heads, each with a hidden dimension of 4096 and an output dimension of 256. 
All heads use intermediate batch normalization and ReLU activations. 
For MoCo v3, we additionally apply batch normalization on the outputs of both heads, following the official implementation. 
We pre-train the baselines for 200 epochs using the LARS optimizer, a linearly scaled learning rate (base value 0.4), a cosine decay schedule applied per training step, and a 10-epoch warm-up. For both methods, the target network is updated using an EMA of the online networks, with a momentum coefficient following a cosine schedule from 0.996 to 1.0. 
All experiments use a batch size of 1024. 

\paragraph{SimSiam baseline.} For SimSiam, we again follow the original training recipe. The projection layer is a three-layer MLP with hidden and output dimensions of 2048, and the prediction head is a two-layer bottleneck MLP with a hidden dimension of 512 and an output dimension of 2048. 
Both heads use intermediate batch normalization and ReLU activations, and the projection head additionally uses batch normalization on its output. 
We pre-train the baseline for 200 epochs with the SGD optimizer, a linearly scaled learning rate (base value 0.05), a batch size of 512, a cosine decay schedule applied per epoch, and no warmup. The prediction head uses a constant learning rate, as in the original paper. 

\paragraph{Base data augmentations.} We adopt the official data augmentation strategies from the respective methods. Unless stated otherwise, training views are obtained by applying the following sequence of transformations: 
\begin{enumerate}
    \item random resized cropping: a random patch of the image is selected, with an area uniformly sampled between 8\% and 100\% of that of the original image, and an aspect ratio logarithmically sampled between 3/4 and 4/3. For SimSiam and MoCo v3, the minimum area is set to 20\%.
    \item random horizontal flipping with a probability of 50\%.
    \item random color jitter: brightness, contrast, saturation, and hue are perturbed with random offsets uniformly sampled for each image. BYOL and MoCo v3 use the ranges $(0.4, 0.4, 0.2, 0.1)$, while SimSiam uses $(0.4, 0.4, 0.4, 0.1)$.
    \item random grayscale with a probability of 20\%. 
    \item random Gaussian blur: the image is blurred with a Gaussian blur kernel of size 23 and a standard deviation uniformly sampled in $[0.1, 2]$. In BYOL and MoCo v3, the transformation is applied with 100\% probability for the first view and 10\% for the second; in SimSiam, both views use a 50\% probability. 
    \item random solarization with probability 20\% in the second view, for BYOL and MoCo v3. 
    \item color normalization: finally, we normalize the color channels by subtracting the per-channel mean and dividing by the per-channel standard deviation estimated on the ImageNet training set.
\end{enumerate}

\paragraph{Multi-predictor multi-crop and multi-task strategies.} All multi-crop and multi-task experiments reuse the hyperparameters of their respective 2-view baselines to ensure a strictly controlled comparison. The multi-crop experiments use two global views and four local views, and the multi-task runs use two global, two local, and one cutout view. Following previous works \cite{caron2020swav,caron2021dino}, local crops have a resolution of 96$\times$96, with the random crop area sampled in the interval $[0.08, 0.25]$.  Global and cutout crops have a resolution of 224$\times$224, with the random crop area sampled in the interval $[0.25, 1.0]$. The two global views follow BYOL's base data augmentation setting described above. For local and cutout views, we apply random horizontal flipping with probability 50\%, color jitter with BYOL's parameter ranges, random grayscale with probability 20\%, and Gaussian blur with probability 50\%. 

\paragraph{Random cutout.} Our random cutout implementation follows the hyperparameter sampling strategy of Torchvision's \texttt{RandomResizedCrop}. We uniformly sample a cutout area between 20\% and 40\% of the original image area, and an aspect ratio logarithmically sampled between 3/4 and 4/3. The selected region is masked with a constant fill value equal to the ImageNet mean color, so that the masked pixels have an expected value of zero after color normalization.

\paragraph{Ablation Study on Data Augmentations.} In the ablation on data augmentations, all experiments reuse the optimization hyperparameters of the 200-epoch BYOL baseline. We vary only the set of augmentations applied during pre-training.

\paragraph{kNN evaluation.} For kNN evaluation, we follow the standard protocol used in prior work on self-supervised learning. Each image is resized to a shorter side of 256 pixels and then center-cropped to $224 \times 224$ before feature extraction. 
For each validation sample, we retrieve its top-k nearest neighbors in the feature space using cosine similarity, and predict the label by a simple majority vote among them. We report the best results across $k=10$ and $k=20$.
Similar to DINO, we observe that with vision transformers, the target network leads to slightly higher accuracy than the online network. 

\paragraph{Linear evaluation.} For linear probing, a linear classifier is trained on top of the frozen backbone features. To eliminate the need for per-model learning rate tuning, feature vectors are standardized using mean and variance statistics computed on ImageNet.
For ResNet 50 architectures, features are extracted after the global average pooling layer, before the projection head. For Vision Transformers (ViT), the extraction protocol follows DINO~\cite{caron2021dino}. For ViT-S, features consist of the concatenated [CLS] tokens from the final four layers. For ViT-B, features are formed by concatenating the [CLS] token with the global average-pooled output patch tokens.
We train the linear head for 100 epochs with the SGD optimizer, a learning rate of 0.005, a batch size of 512, no weight decay, and a cosine learning rate schedule. 
During training, we apply random horizontal flipping with probability 50\% and random cropping that keeps at least 8\% of the image area, followed by resizing to $224 \times 224$ pixels. At evaluation time, images are resized so that the shorter side is 256 pixels, then center-cropped to $224 \times 224$ pixels.

\section{Additional Evaluations} \label{app:additional_evals}

\begin{table}[tb]
\centering\small
\caption{Semi-supervised training on ImageNet-1k with 1\% and 10\% of labels. We report top-1 and top-5 validation set accuracies. $^\star$We run the DINO semi-supervised evaluation under the same procedure.}
\label{table:semisup}
\begin{tabular}{lcccc}
    \toprule
    \multirow{2}{*}{Method} & \multicolumn{2}{c}{ Top-1 } & \multicolumn{2}{c}{ Top-5 } \\
    & $1 \%$ & $10 \%$ & $1 \%$ & $10 \%$ \\
    \midrule
    {\color{gray} supervised} & {\color{gray} 25.4} & {\color{gray} 56.4} & {\color{gray} 48.4} & {\color{gray} 80.4} \\
    SimCLR & 48.3 & 65.6 & 75.5 & 87.8 \\
    BYOL  & 53.2 & 68.8 & 78.4 & 89.0 \\
    DINO$^\star$ & 49.1 & 69.0 & 76.1 & 89.7 \\
    SwAV & 53.9 & 70.2 & 78.5 & 89.9 \\
    Barlow Twins & 55.0 & 69.7 & 79.2 & 89.3 \\
    NNCLR & 56.4 & 69.8 & 80.7 & 89.3 \\
    C-BYOL & \textbf{60.6} & 70.5 & 83.4 & 90.0 \\
    ReLIC v2 & 58.1 & 72.4 & 81.3 & 91.2 \\
    \rowcolor{lightblue}
    MULAN & 60.4 & \textbf{72.5} & \textbf{84.0} & \textbf{91.3} \\ %
    \bottomrule
\end{tabular}
\end{table}

\paragraph{Semi-supervised learning.} We follow the semi-supervised learning protocol from \cite{zhai2019s4l,chen2020simclr,grill2020byol}, which we describe next for completeness. We fine-tune the pre-trained ResNet-50 backbone from \Cref{sec:sota_results} using subsets of ImageNet-1k labels (1\% and 10\%), as provided by \cite{chen2020simclr}. During training,  we apply the same data augmentations as for linear evaluation: random horizontal flipping and random cropping (with scales in the range 0.08-1), followed by resizing to 224$\times$224 pixels. At test time, images are resized to 256 pixels along the shorter side using bicubic resampling, followed by a center crop to 224$\times$224 pixels. In all cases, we use the same color normalization as during pre-training. We attach a linear classification head to the backbone and train the network with a softmax cross-entropy loss using SGD with Nesterov momentum 0.9, a batch size of 1024, and no additional regularization (e.g., no weight decay). We sweep over learning rate values in $\left\{0.005, 0.01, 0.02, 0.05, 0.1 \right\}$ and training schedules of 30 and 50 epochs, and report the test accuracy of the best configuration in \Cref{table:semisup}. 

Our approach significantly outperforms BYOL and achieves state-of-the-art performance. In the $1\%$ setting, it outperforms ReLIC v2 by a large margin and matches the performance of C-BYOL \cite{lee2021cbyol}, a compressive variant of BYOL. Notably, the C-BYOL objective could in principle be combined with our augmentation strategy, which we leave for future work.

\paragraph{Fine-tuning.} We also fine-tune the pre-trained models on the full ImageNet-1k training set using the same simple protocol, with only two data augmentations and no explicit regularization. As reported in \Cref{table:finetune}, MULAN improves over standard BYOL by +1.6 points in top-1 accuracy and by +1.1 points in top-5 accuracy.

\begin{table}[htb]
\centering\small
\caption{Fine-tuning on ImageNet-1k with neither heavy data augmentations nor regularization.}
\label{table:finetune}
\begin{tabular}{lcc}
    \toprule
    Method & Top-1 & Top-5 \\
    \midrule
    SimCLR & 76.0 & 93.1 \\
    BYOL & 77.7 & 93.7 \\
    \rowcolor{lightblue} MULAN & \textbf{79.3} & \textbf{94.8} \\
    \bottomrule %
\end{tabular}
\end{table}

\section{Timing Analysis} \label{app:timing}

\Cref{tab:efficiency} compares training cost and wall-clock efficiency across different BYOL training strategies. Although multi-task training increases the per-epoch cost, it is much more wall-clock efficient. The 200-epoch multi-task model (72h) outperforms the 1000-epoch baseline (219h) by 1.3\% accuracy, corresponding to an approximately $3\times$ gain in training efficiency. 

Furthermore, the method is implemented with per-view forward and backward passes, keeping memory usage independent of the number of views. This reordering is mathematically equivalent to the standard multi-view formulation, yielding identical gradients. Under this setup, peak per-GPU memory consumption is 12.7 GB for all three augmentation strategies.

\begin{table}[htb]
\caption{Training efficiency on BYOL (ResNet-50) using a single node with 4$\times$ Nvidia A100 (80GB) GPUs.}
\label{tab:efficiency}
\centering \small
\begin{tabular}{lrrrrr}
    \toprule
    Strategy & Epochs & Time/Ep & Total (h) & Lin. \\
    \midrule
    baseline & 200 & 13:09 & 44 & 70.7 \\
    multi-pred m-c & 200 & 19:51 & 66 & 74.7 \\
    multi-task & 200 & 21:35 & 72 & \textbf{75.6} \\
    baseline & 1000 & 13:09 & 219 & 74.3 \\
    \bottomrule
\end{tabular}
\end{table}

\section{Extended Analysis of Multi-Crop Instability} \label{app:extended_mc_analysis}

To understand multi-crop instability in self-predictive methods, we combined insights from prior work with observations from our experiments. Caron \etal~\cite{caron2021dino} reported that BYOL with multi-crop suffers a significant accuracy drop for both ResNet-50 and ViT-S backbones, and Morningstar \etal~\cite{morningstar2024augsvsalgos} observed analogous drops for MoCo~v3. Similar instability occurs in SimSiam~\cite{chen2021simsiam}, which does not use an EMA target. Together, these observations indicate that neither the EMA update nor the backbone architecture alone can explain the instability. 

We also tested simple mitigation strategies reported in the literature, such as reducing batch size or learning rate. While these adjustments can prevent collapse, they do not restore performance to the level of the standard two-view baseline. This suggests that the optimization scale or rate is not the primary cause.

Taken together, these analyses support the conclusion that multi-crop instability is primarily associated with the predictor being asked to solve heterogeneous latent-space tasks, motivating the decoupled predictor solution described in the main text.

\section{On Training Self-predictive Vision Transformers} \label{app:vit_details}

This section details our training strategy for vision transformers, intending to facilitate the tuning of future methods. 

We first adopt two changes relative to the ResNet setting: a deeper projection head following MoCo v3, and stochastic depth \cite{huang2016stoch_depth} applied to the online network with a dropout probability of 10\%, following DINO.

\begin{figure}[ht]
    \centering
    \def\svgwidth{0.84\textwidth}
    {\footnotesize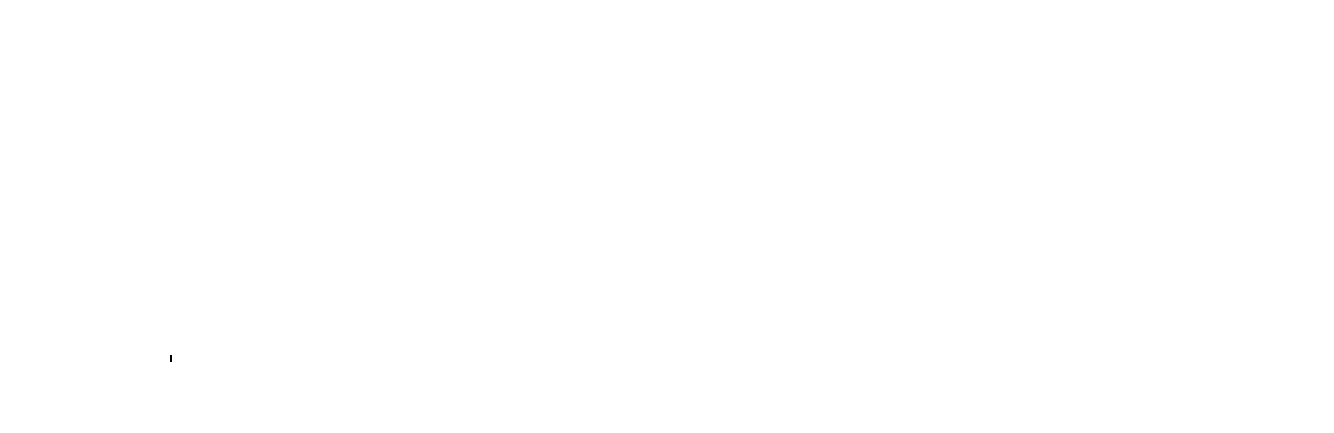}
    \caption{Training loss curves for MULAN with a ViT-B backbone on ImageNet (first 80k steps shown). We compare AdamW with $\beta_2=0.999$ and no gradient clipping against $\beta_2=0.98$ with gradient clipping at $0.5$. The latter approach avoids loss spikes and achieves higher final accuracy.}
    \label{fig:beta2_impact_vitb}
\end{figure}

Training vision transformers is known to require carefully tuned schedules and hyperparameters, particularly in self-supervised settings \cite{touvron2021deit,chen2021mocov3,caron2021dino}. A notable challenge is training instability: MoCo v3 \cite{chen2021mocov3} reported sudden loss spikes that degrade performance and prevent the use of large learning rates, and showed that freezing the patch projection layer mitigates this issue across several methods (MoCo v3, BYOL, and SimCLR). In our multi-task setting, however, freezing the first layer leads to lower accuracy. Instead, we find that lowering the Adam optimizer's $\beta_2$ parameter (the exponential moving average of squared gradients) from 0.999 to 0.98, combined with gradient clipping, effectively mitigates these instabilities (\Cref{fig:beta2_impact_vitb}) and unlocks substantial performance gains. This observation is consistent with findings from the natural language processing literature \cite{orvieto2026adam_secret,bai2025beta2_spikes,brown2020gpt3,touvron2023llama}.

A second challenge is that self-predictive methods tend to saturate early when trained with vision transformers. This is particularly pronounced for BYOL, which does not benefit from training schedules longer than 300 epochs \cite{caron2021dino}, and our multi-task variant saturates even earlier, around epoch 200. Notably, our method already matches DINO's 800-epoch accuracy at 200 epochs, highlighting its potential. Enabling it to benefit from longer schedules, whether through regularization or refined hyperparameter tuning, remains an open challenge and a key direction for future work.

\section{Negative Results and Future Directions} \label{app:negative_results}

While augmenting the multi-crop strategy with asymmetric cutout views yielded significant gains (\Cref{tab:ablation_views}), several alternative spatial and mixing transformations proved ineffective. To guide the development of future work, we detail the online view-generation strategies that failed to improve performance: 
\begin{itemize}
    \item \textbf{Random Rotation}: Applying stochastic rotations to the input image before cropping. 
    \item \textbf{Patch Shuffling}: Dividing the input into a grid of $4 \times 4$ patches (similar to Vision Transformer tokenization) and performing a random shuffle.
    \item \textbf{CutMix-like}: Adopting a CutMix \cite{yun2019cutmix} approach where the online branch processes a blended image. The corresponding target was defined as a weighted average of the individual representations (passed separately to the target branch). 
\end{itemize}

\begin{table}[htb]
\caption{Alternative pre-training strategies on BYOL (ResNet-50).}
\label{tab:alternatie_augs}
\centering \small
\begin{tabular}{lrr}
    \toprule
    Pre-training Strategy & kNN & Lin.  \\
    \midrule
    glob. + loc. & 69.5 & 74.7 \\
    glob. + loc. + \textbf{rotate} & 68.9 & 75.1 \\
    glob. + loc. + \textbf{shuffle} & 69.7 & 75.1 \\
    glob. + loc. + \textbf{cutout} & 69.9 & 75.6 \\
    glob. + loc. + \textbf{shuffle} + \textbf{cutout} & 70.0 & 75.7 \\
    \bottomrule
\end{tabular}
\end{table}
The first two strategies yielded small gains (\Cref{tab:alternatie_augs}) that did not justify the increased computational overhead, while the third degraded performance. 
All three approaches failed to yield meaningful representations in the standalone validation setting of \Cref{sec:assym_cutout_validation}, confirming that this setting is a reliable prerequisite before integrating new tasks into the multi-task formulation.

An alternative approach for \textit{Random Rotation} or \textit{Patch Shuffling} would be to provide a conditioning signal to the predictors (e.g., the specific rotation angle). This modification would transform the method from a standard Siamese or Joint-Embedding Architecture into a Joint-Embedding Predictive Architecture, shifting the focus from learning invariant features to task-conditional prediction and enabling richer forms of self-supervision.

%% file: figures/beta2_comparison.pdf_tex
\begingroup%
  \makeatletter%
  \providecommand\color[2][]{%
    \errmessage{(Inkscape) Color is used for the text in Inkscape, but the package 'color.sty' is not loaded}%
    \renewcommand\color[2][]{}%
  }%
  \providecommand\transparent[1]{%
    \errmessage{(Inkscape) Transparency is used (non-zero) for the text in Inkscape, but the package 'transparent.sty' is not loaded}%
    \renewcommand\transparent[1]{}%
  }%
  \providecommand\rotatebox[2]{#2}%
  \newcommand*\fsize{\dimexpr\f@size pt\relax}%
  \newcommand*\lineheight[1]{\fontsize{\fsize}{#1\fsize}\selectfont}%
  \ifx\svgwidth\undefined%
    \setlength{\unitlength}{640.158125bp}%
    \ifx\svgscale\undefined%
      \relax%
    \else%
      \setlength{\unitlength}{\unitlength * \real{\svgscale}}%
    \fi%
  \else%
    \setlength{\unitlength}{\svgwidth}%
  \fi%
  \global\let\svgwidth\undefined%
  \global\let\svgscale\undefined%
  \makeatother%
  \begin{picture}(1,0.32466271)%
    \lineheight{1}%
    \setlength\tabcolsep{0pt}%
    \put(0,0){\includegraphics[width=\unitlength,page=1]{beta2_comparison.pdf}}%
    \put(0.12832641,0.03586272){\makebox(0,0)[t]{\lineheight{1.25}\smash{\begin{tabular}[t]{c}0k\end{tabular}}}}%
    \put(0,0){\includegraphics[width=\unitlength,page=2]{beta2_comparison.pdf}}%
    \put(0.23407198,0.03586272){\makebox(0,0)[t]{\lineheight{1.25}\smash{\begin{tabular}[t]{c}10k\end{tabular}}}}%
    \put(0,0){\includegraphics[width=\unitlength,page=3]{beta2_comparison.pdf}}%
    \put(0.33981756,0.03586272){\makebox(0,0)[t]{\lineheight{1.25}\smash{\begin{tabular}[t]{c}20k\end{tabular}}}}%
    \put(0,0){\includegraphics[width=\unitlength,page=4]{beta2_comparison.pdf}}%
    \put(0.44556314,0.03586272){\makebox(0,0)[t]{\lineheight{1.25}\smash{\begin{tabular}[t]{c}30k\end{tabular}}}}%
    \put(0,0){\includegraphics[width=\unitlength,page=5]{beta2_comparison.pdf}}%
    \put(0.5513087,0.03586272){\makebox(0,0)[t]{\lineheight{1.25}\smash{\begin{tabular}[t]{c}40k\end{tabular}}}}%
    \put(0,0){\includegraphics[width=\unitlength,page=6]{beta2_comparison.pdf}}%
    \put(0.65705431,0.03586272){\makebox(0,0)[t]{\lineheight{1.25}\smash{\begin{tabular}[t]{c}50k\end{tabular}}}}%
    \put(0,0){\includegraphics[width=\unitlength,page=7]{beta2_comparison.pdf}}%
    \put(0.76279986,0.03586272){\makebox(0,0)[t]{\lineheight{1.25}\smash{\begin{tabular}[t]{c}60k\end{tabular}}}}%
    \put(0,0){\includegraphics[width=\unitlength,page=8]{beta2_comparison.pdf}}%
    \put(0.86854542,0.03586272){\makebox(0,0)[t]{\lineheight{1.25}\smash{\begin{tabular}[t]{c}70k\end{tabular}}}}%
    \put(0,0){\includegraphics[width=\unitlength,page=9]{beta2_comparison.pdf}}%
    \put(0.97429103,0.03586272){\makebox(0,0)[t]{\lineheight{1.25}\smash{\begin{tabular}[t]{c}80k\end{tabular}}}}%
    \put(0.55025124,0.01449592){\makebox(0,0)[t]{\lineheight{1.25}\smash{\begin{tabular}[t]{c}Training Step\end{tabular}}}}%
    \put(0,0){\includegraphics[width=\unitlength,page=10]{beta2_comparison.pdf}}%
    \put(0.11527669,0.11030236){\makebox(0,0)[rt]{\lineheight{1.25}\smash{\begin{tabular}[t]{r}$10^{0}$\end{tabular}}}}%
    \put(0,0){\includegraphics[width=\unitlength,page=11]{beta2_comparison.pdf}}%
    \put(0.11527669,0.30154586){\makebox(0,0)[rt]{\lineheight{1.25}\smash{\begin{tabular}[t]{r}$10^{1}$\end{tabular}}}}%
    \put(0,0){\includegraphics[width=\unitlength,page=12]{beta2_comparison.pdf}}%
    \put(0.0621697,0.18307391){\rotatebox{90}{\makebox(0,0)[t]{\lineheight{1.25}\smash{\begin{tabular}[t]{c}Loss\end{tabular}}}}}%
    \put(0,0){\includegraphics[width=\unitlength,page=13]{beta2_comparison.pdf}}%
    \put(0.82430615,0.28155202){\makebox(0,0)[lt]{\lineheight{1.25}\smash{\begin{tabular}[t]{l}$\beta_2=0.999$\end{tabular}}}}%
    \put(0,0){\includegraphics[width=\unitlength,page=14]{beta2_comparison.pdf}}%
    \put(0.82430615,0.25818865){\makebox(0,0)[lt]{\lineheight{1.25}\smash{\begin{tabular}[t]{l}$\beta_2=0.98$\end{tabular}}}}%
  \end{picture}%
\endgroup%